\pdfoutput=1

\documentclass[11pt]{article}

\usepackage{acl}

\usepackage{times}
\usepackage{latexsym}
\usepackage[algoruled,ruled,vlined,noend]{algorithm2e}
\usepackage{amsmath}
\usepackage{times}
\usepackage{latexsym}
\usepackage{pgfplots}
\usepackage{graphicx}
\usepackage{enumitem}
\usepackage{xcolor,colortbl}  
\usepackage{tikz}
\usepackage{diagbox}
\usepackage{tkz-tab}
\usepackage{caption}
\usepackage{booktabs}
\usepackage{latexsym}
\usepackage{amssymb}
\usepackage{amsmath}
\usepackage{subcaption}
\usepackage{natbib}
\usepackage{blindtext}
\usepackage{url}
\usepackage{color}

\usepackage{url}
\usepackage{hyperref}    
\usepackage{cleveref}
\SetAlFnt{\small}
\SetAlCapFnt{\small}
\SetAlCapNameFnt{\small}
\usepackage{ntheorem}

\usepackage{booktabs, tabularx}
\usepackage{tfrupee}  
\usepackage{caption}

\usepackage[colorinlistoftodos]{todonotes}

\usepackage[T1]{fontenc}
\usepackage[utf8]{inputenc}

\usepackage{microtype}

\title{Finding Memo:\\ Extractive Memorization in Constrained Sequence Generation Tasks}

\author{Vikas Raunak\qquad Arul Menezes\\
Microsoft Azure AI \\
Redmond, Washington \\
\texttt{\{viraunak,arulm\}@microsoft.com}}

\begin{document}
\maketitle
\begin{abstract}


Memorization presents a challenge for several constrained Natural Language Generation (NLG) tasks such as Neural Machine Translation (NMT), wherein the proclivity of neural models to memorize  noisy and atypical samples reacts adversely with the noisy (web crawled) datasets. However, previous studies of memorization in constrained NLG tasks have only focused on counterfactual memorization, linking it to the problem of hallucinations. In this work, we propose a new, inexpensive algorithm for extractive memorization (exact training data generation under insufficient context) in constrained sequence generation tasks and use it to study extractive memorization and its effects in NMT. We demonstrate that extractive memorization poses a serious threat to NMT reliability by qualitatively and quantitatively characterizing the memorized samples as well as the model behavior in their vicinity. Based on empirical observations, we develop a simple algorithm which elicits non-memorized translations of memorized samples from the same model, for a large fraction of such samples. Finally, we show that the proposed algorithm could also be leveraged to mitigate memorization in the model through finetuning. We have released the code to reproduce our results at \href{https://github.com/vyraun/Finding-Memo}{https://github.com/vyraun/Finding-Memo}.

\end{abstract}

\section{Introduction}
\label{sec:introduction}

Previous studies \cite{arpit_memorization, feldman_long_tail, rethinking_generalization_2} have shown that neural networks capture regular patterns in the training data (generalization) while simultaneously fitting noisy and atypical samples using brute-force (memorization). For constrained Natural Language Generation tasks such as Neural Machine Translation (NMT), which rely heavily on noisy (web crawled) data for training high-capacity neural networks, this creates an inherent reliability problem. For example, memorizations could manifest themselves in the form of catastrophic translation errors on specific samples despite high average model performance \cite{raunak_nmt_hal}. It is also \textit{likely} that the memorization of a specific sample could corrupt the translations of samples in its vicinity. Therefore, exploring, quantifying and alleviating the impact of memorization is of critical importance for improving the reliability of such systems.

Yet, most of the work on memorization in natural language processing (NLP) has focused either on classification \cite{jiang_mem} or on unconstrained generation tasks, predominantly language modeling \cite{carlini_lm_extraction, zhang_counterfactual, bpe_memorization, palm, fb_memorization, bert_memorisation, mor_geva}. In this work, we fill a gap in the literature by developing an analogue of extractive memorization for constrained sequence generation tasks in general and NMT in particular. Our main contributions are:
\vspace{-0.4em}

\begin{enumerate}
\itemsep-0.0em 
    \item We propose a new, inexpensive algorithm for studying extractive memorization in constrained sequence generation tasks and use it to characterize memorization in NMT. 
    \item We demonstrate that extractive memorization poses a serious threat to NMT reliability by quantitatively and qualitatively analyzing the memorized samples and the neighborhood effects of such memorization. We also demonstrate that the memorized instances could be used to generate errors in \textit{disparate} systems.
    \item Based on an analysis of the neighborhood effects of memorization, we develop a simple memorization mitigation algorithm which produces non-memorized (higher quality) outputs for a large fraction of memorized samples. 
    \item We show that the outputs produced by the memorization mitigation algorithm could also be used to directly impart corrective behavior into the model through finetuning.
\end{enumerate}

\begin{table*}
\centering
\scalebox{0.90}{
\begin{tabular}{c|c|c|c||c|c|c}
\hline
\textbf{Repetitions} & \textbf{Total Samples}    & \textbf{Memorized}    & \textbf{Ratio (\%)}  & \textbf{Perturb Prefix} & \textbf{Perturb Suffix} & \textbf{Perturb Start} \\ \hline
1    & 100,000  & 174 & 0.17 & 17.58 \% & 43.24 \% & 12.29 \% \\ 
2    & 100,000 & 317 & 0.32 & 11.67 \% & 62.84 \% & 4.98 \% \\
3    & 5,381 & 17 & 0.32 & 28.42 \% & 49.52 \% & 18.82 \% \\
4    & 1,885 & 5 & 0.26 & 27.40 \% & 34.00 \% & 8.00 \% \\
5    & 976 & 7 & 0.72 & 26.67 \% & 70.00 \% & 11.42 \%  \\ \hline
\textbf{1-5}  & \textbf{208,242} & \textbf{ 520} & \textbf{0.25} & \textbf{16.65 \%} & \textbf{51.65 \%} & \textbf{8.00 \%}  \\ \hline
\end{tabular}}
\caption{\textbf{Quantifying Extractive Memorization:} Number of Memorized Samples (using Algorithm \ref{algo:algo_1}) and Neighborhood Effects of Memorization (using Algorithm \ref{algo:algo_2}) across different training data frequency buckets.}
\label{tab:main_table}
\vspace{-1.00em}
\end{table*}

\section{Related Work}
\label{sec:related_work}

Our work is concerned with the phenomenon of memorization in constrained natural language generation in general and NMT in particular. The main challenge in analyzing memorization is to determine which samples have been memorized by the model during training. There exist two key algorithms to elicit memorized samples, each yielding a distinctive operational definition of memorization:

\begin{enumerate}
    \item \textbf{Counterfactual Memorization}: \citet{feldman_operator} study label memorization and propose to estimate the memorization value of a training sample by training multiple models on different random subsets of the training data and then measuring the deviation in the sample's classification accuracy under inclusion/exclusion. This definition of memorization was further extended to arbitrary performance measures by \citet{raunak_nmt_hal} to study memorization in NMT and by \citet{zhang_counterfactual} to study memorization in language models. However, a practical limitation of analysis based on this definition is the prohibitive computational cost (multiple model trainings) associated with computing memorization values for each training sample. 
    \item \textbf{Extractive Memorization}: \citet{carlini_lm_extraction} propose a data-extraction based definition of memorization to study memorization in language models. Therein, a training string $s$ is extractable if there exists a prefix $c$ that could exactly generate $s$ under an appropriate sampling strategy (e.g. greedy decoding). This definition has the benefit of being computationally inexpensive, although it doesn't have any existing analogue for constrained natural language generation tasks such as NMT.
\end{enumerate}

In the next section, we define extractive memorization for constrained sequence generation tasks and apply it to NMT, in section \ref{sec:neighborhood_effect} we estimate the neighborhood effect of such memorizations and in section \ref{sec:memorization_recovery} we propose a simple algorithm for recovering correct translations of memorized samples. 

\section{Extractive Memorization}
\label{sec:extractive_memorization}

We present our definition of extractive memorization as Algorithm \ref{algo:algo_1}. Analogous to extractive memorization in language models \cite{carlini_lm_extraction}, this definition labels an input sentence (source) as being memorized if its transduction (translation) could be replicated exactly with a prefix considerably shorter than the length of the full input sentence (source), under greedy decoding. Operationally, we set prefix ratio threshold ($p$) to $0.75$.

\begin{algorithm}[ht]
 \SetAlgoNoLine
 \SetNoFillComment
 \KwData{Trained NMT Model $T$, Training Dataset $S$, Prefix Ratio Threshold $p$}
 \KwResult{Memorized Samples $M$, Prefix Lengths $L$} 
 Greedily Translate Sources in $S$ using $T$\;
 $M_{1}$ = Sources with translations matching References\;
 Greedily Translate Prefixes of Sources in $M_{1}$ using $T$\;
  $M_{2}$ = Sources with Prefixes producing References\;
  \For{$M_{2}^{i}$ in $M_{2}$}{
     $n$ = Length of the Source $M_{2}^{i}$ \;
     $l$ = Length of Smallest Prefix producing the Ref\;
     \If{ $ \frac{l}{n} \leq p $ }
     {Add $M_{2}^{i}$ to $M$ and Add $l$ to $L$\;}
  }
\caption{Extractive Memorization in NMT\label{algo:algo_1}}
\end{algorithm}

Next, we apply this definition of memorization on a strong Transformer-Big \cite{transformer} baseline trained on the 48.2M WMT20 En-De parallel corpus \cite{wmt_2020}. We describe the dataset, model and training details in Appendix A.

\begin{table*}[ht]
    \begin{tabularx}{\linewidth}{ l l X} 
        \toprule
    \textbf{Provenance}  & \textbf{Source} & \textbf{Translation}\\
        \midrule 
Training Data & Why study in Peru? Spanish Courses & Warum in Peru studieren? \\ \midrule        
Perturb Suffix & Why study in Peru? University Courses &  Warum in Peru studieren? \\
Perturb Suffix & Why study in Peru? Short Courses &  Warum in Peru studieren? \\
Perturb Suffix & Why study in Peru? Summer Courses &  Warum in Peru studieren? \\ \midrule
Perturb Prefix & You study in Peru? Spanish Courses &  Sie studieren in Peru? Spanischkurse \\
Perturb Prefix & Advanced study in Peru? Spanish Courses &  Weiterbildung in Peru? Spanischkurse \\
        \bottomrule
    \end{tabularx}
    \caption{\textbf{Memorization Example}: An example to illustrate the phenomenon of Memorization (elicited using Algorithm \ref{algo:algo_1}) and the ensuing Neighborhood Effect of such memorization (measured using Algorithm \ref{algo:algo_2}). This WMT20 English to German training sample is memorized with a prefix length ratio of 0.67 ($\leq$ 0.75).}
    \label{tab:memorization_example}
    \vspace{-0.8em}
\end{table*}

Qualitatively, we observe that the memorized samples detected by Algorithm \ref{algo:algo_1} mostly consist of low-quality samples -- templatized source sentences and noisy translations. To analyze the results quantitatively, similar to \citet{carlini_lm_analysis}, we bucket the training data pairs in terms of their repetitions in the training data. Owing to the sparsity of data with greater than 5 repetitions we report results in the range of 1-5 repetitions. Further, for repetition values 1 and 2, we select 100K random samples for analysis. We observe two key results:

\textbf{Repetitions vs Memorization:}  Table \ref{tab:main_table} shows that the percentage of samples memorized is higher for repeated training samples, compared to samples present only once in the training data, with a Pearson's correlation coefficient of 0.778 between the number of repetitions and memorizations.

\textbf{Quality of Memorized Samples:} Table \ref{tab:memorized_set_quality} shows that both the quality of memorized samples, measured using COMET-QE \cite{comet}, a state-of-the-art Quality Estimation model for MT as well as their lexical diversity, measured using Type-to-Token Ratio (TTR) \cite{ttr_mt, gemv2} is worse when compared to the total samples.

\vspace{-0.2em}
\begin{table}[!htbp]
\centering
\scalebox{0.90}{
\begin{tabular}{c|c|c|c|c}
\hline 
\textbf{Rep.} & \textbf{T-COM} & \textbf{M-COM} & \textbf{T-TTR} & \textbf{M-TTR} \\ \hline
1   & 42.91  & 32.81 & 9.81 & 44.87 \\ 
2   & 60.63  & 62.56 & 8.85 & 36.92 \\ 
3   & 73.19  & 37.31 & 20.03 & 68.97\\ 
4   & 70.18  & 61.92 & 27.72 & 78.85 \\ 
5   & 68.41  & 46.65 & 33.90 & 73.47\\ \hline
\textbf{1-5}   & \textbf{54.79}  & \textbf{51.57} & \textbf{7.64} & \textbf{38.72} \\ \hline
\end{tabular}}
\caption{Comparison of \textbf{COM}ET-QE $\uparrow$ and \textbf{TTR} $\downarrow$ for Memorized (\textbf{M}) Samples vs Total (\textbf{T}) Samples.}
\label{tab:memorized_set_quality}
\vspace{-0.6em}
\end{table}

The above two results are similar to the results in language modeling \cite{carlini_lm_analysis} and serve to demonstrate the utility of Algorithm \ref{algo:algo_1} in analyzing extractive memorization in NMT.

\section{Neighborhood Effect of Memorization}
\label{sec:neighborhood_effect}

To measure the neighborhood effect of memorization, we generate new source sentences in the vicinity of the memorized samples through perturbations and test whether they generate the same output under greedy decoding. Table \ref{tab:memorization_example} shows a memorized training sample from Section \ref{sec:extractive_memorization}, alongside translations generated from perturbations at different positions in the source. The perturbations (substitutions) were generated using BERT-Cased \cite{bert}. We define suffix positions and prefix positions based on the recorded prefix lengths ($L$) in Algorithm \ref{algo:algo_1}. Specifically, we use Algorithm \ref{algo:algo_2}, which generates new sources in the neighborhood of a memorized input source by perturbing its tokens at prefix ($P$), suffix ($S$) or the beginning ($B$) positions and then computes how many such new sources still translate to the same memorized output. The resulting effect measure $N$ is higher if the sources still produce the same memorized output under perturbations at different positions. The intuition behind Algorithm 2 is that we want to explore meaningful source sequences around the memorized source sentence. Therefore, we make changes to only a single position at a time, to keep the generated sequence close to the original sequence and substitute that position with a word that fits well in the particular context. Our `neighbour' definition is loosely inspired by the differential privacy literature \cite{dwork2014algorithmic}, where a neighbour is used to mean datasets differing in only one item.

\begin{algorithm}[ht]
 \SetAlgoNoLine
 \SetNoFillComment
 \KwData{Memorized Samples $M$, Prefix Lengths $L$, Masked Language Model $W$, Candidates $K$ }
 \KwResult{Perturbed Sources $S^{'}$, Effect Measure $N$}

 \For{$M_{i}$, $L_{i}$ in $M$, $L$ }{
 $n$ = Length of the Source $M_{i}$\;
 $P_{i}$ = [1,...,$L_{i}-1$],  $S_{i}$ = [$L_{i}+1$,..., $n$], $B_{i}$ = [1]\;
  \For{$j$ in $P_{i}$, $S_{i}$, $B_{i}$ }{
 Generate $K$ substitutes at $M_{i}[j]$ using $W$\;
 Translate each new Source and Add to $S^{'}$\;
 }
 }
$N$ = Fraction of $S^{'}$ Generating Memorized Outputs\;
\caption{ Estimating Neighborhood Effect \label{algo:algo_2}}
\vspace{-0.1em}
\end{algorithm}

\textbf{Results}: The last 3 columns of Table \ref{tab:main_table} present the results of applying Algorithm \ref{algo:algo_2} on memorized sources ($K$=5). We find that considerable percentages of perturbations still yield the same translations, and that this effect is highly position-dependent. For example, for the unique memorized samples (repetitions = 1), perturbing the suffix produces no changes in translations for 43.2\% of the new sources; while for prefix perturbations the same memorized output is produced only 17.6\%. Further, perturbing the first token of the source is highly successful in generating a different (non-memorized) output. The results demonstrate that a single memorization may therefore be able to corrupt the translations of multiple input sequences in its vicinity.

Table \ref{tab:main_table} and \ref{tab:memorization_example} show that while the translations of memorized sources frequently remain invariant under suffix perturbations, translations under prefix-perturbations do change more frequently. We hypothesize that this results from the model switching away from its memorization mode, owing to a change in the input prefix used for memorization. 

\section{Generating Non-Memorized Outputs}
\label{sec:memorization_recovery}

The results in section \ref{sec:neighborhood_effect} (and table \ref{tab:memorization_example}) show that the model is able to generate non-memorized translations quite frequently if memorized source's start token is perturbed. We demonstrate that this fact could be exploited to \textit{recover} the \textit{non-memorized} translations of a memorized sample from the same model with a surprisingly high-frequency. We call this task Memorization-Mitigation and present a simple technique to do so in Algorithm \ref{algo:algo_3}.

\begin{algorithm}[ht]
 \SetAlgoNoLine
 \SetNoFillComment
 \KwData{Trained NMT Model $T$, Memorized Samples $M$, Recovery Symbol $X$}
 \KwResult{Memorized Samples with New Outputs $F$}

 \For{$S_{i}$ in $M$}{
  $R_{i}$ = Translate $S_{i}$ using $T$\;
  $S_{i}^{aug}$ = $X$ + $S_{i}$\;
  $R_{i}^{aug}$ = Translate $S_{i}^{aug}$ with $T$\;
\If{ $R_{i}^{aug} \neq R_{i}$ and  $R_{i}^{aug}$ starts with $X$ }
{ Strip $X$ from $R_{i}^{aug}$, Add ($S_{i}$, $R_{i}^{aug}$ ) to $F$\; }
}
\caption{Generating Non-Memorized Outputs \label{algo:algo_3}}
\end{algorithm}

\vspace{-0.6em}
\begin{figure}[ht]%
\centering%
\fbox{\parbox{\dimexpr\linewidth-2\fboxsep+2.5\fboxrule\relax}{ 
\textbf{Source}: Victor Emmanuel II of Italy  \\
\textbf{Translation (Reference)}: Viktor Emanuel II. \\
\textbf{Perturbed Source}: ! Victor Emmanuel II of Italy  \\
\textbf{Translation}: ! Viktor Emanuel II von Italien \\
\textbf{Stripped Output}: Viktor Emanuel II von Italien
}}%

\caption[]{ \textbf{Generating Non-Memorized Output} The example shows how the inclusion of an `isolated' symbol early in the input prefix elicits a non-memorized translation from the model using Algorithm \ref{algo:algo_3}.}%
\label{fig:recovery_example}%
\end{figure}

Algorithm \ref{algo:algo_3} uses the idea of prefix perturbation to elicit a non-memorized translation from the model. As a perturbation, it appends a symbol $X$ to the source sentence, which is chosen such that it could be translated in isolation, without semantically altering the source. We find that new translations for 65.2\% of the memorized samples in Table \ref{tab:main_table} (339/520) could be recovered using Algorithm \ref{algo:algo_3} with symbol $X$ set to `!'. We find that the results vary with different choices of symbol $X$ and report the best result after trying 5 different symbols. An example of applying Algorithm \ref{algo:algo_3} on a memorized sample is presented in Figure \ref{fig:recovery_example}. 

Further, we find that the new (non-memorized) translations are of much higher quality than the memorized outputs. Table \ref{tab:post_recovery} presents the results by comparing the new translations to the memorized translations using COMET-QE and TTR. 

\begin{table}[!htbp]
\centering
\scalebox{0.90}{
\begin{tabular}{c|c|c}
\hline 
\textbf{Set} & \textbf{COMET-QE $\uparrow$ } & \textbf{TTR $\downarrow$ } \\ \hline
\textbf{Memorized}  & 54.57 & 57.16 \\
\textbf{Non-Memorized}  & \textbf{84.00} & \textbf{40.47} \\ \hline
\end{tabular}}
\caption{\textbf{Quality of Non-Memorized Outputs}: Comparing the Quality of Algorithm 3 outputs against Memorized References using COMET-QE and Char-Ratio.}
\label{tab:post_recovery}
\vspace{-1.5em}
\end{table}

\section{Mitigating Memorization in the Model}
\label{sec:recovery_finetuning}

The previous section demonstrated that the non-memorized translations of memorized samples could be recovered from the \textit{same} model by applying Algorithm \ref{algo:algo_3}. In this section, we investigate whether this corrective behavior could directly be imparted to the model through finetuning using the corpus $F$ obtained by applying Algorithm \ref{algo:algo_3}.

\begin{table}[!htbp]
\centering
\scalebox{0.90}{
\begin{tabular}{c|c|c}
\hline
\textbf{Measurement} & \textbf{Base} & \textbf{Finetuned} \\ \hline
\textbf{WMT20 Test BLEU}  & 32.9 & \textbf{33.5 }\\
\textbf{Memorized COMET-QE}  & 54.57 & \textbf{68.94} \\ \hline
\end{tabular}}
\caption{\textbf{Post-Finetuning Quality}: Comparing Model Quality on WMT20 Test and the Memorized Set.}
\label{tab:post_recovery}
\vspace{-0.5em}
\end{table}

To test this, we finetune the last checkpoint of the trained WMT20 model for one epoch on approximately 10K data pairs, comprising of 10K parallel data samples drawn randomly from the training corpus as well as the corpus \textit{F}. We compare the model prior to and post finetuning, in terms of both general performance on the WMT20 test set with BLEU \cite{bleu, sacrebleu} and on the corpus $F$ with COMET-QE. Table \ref{tab:post_recovery} shows that given the non-memorized translations, such corrective behavior could be imparted to the model without impacting general quality.

\textbf{Further Experiments:} Throughout sections \ref{sec:extractive_memorization}-\ref{sec:recovery_finetuning}, we have used the trained WMT20 En-De system to conduct various experiments exploring extractive memorization. However, the reported phenomena are quite general in nature, observable across different language pairs, systems and datasets; the results for other experiments are presented in Appendix \ref{sec:appendix_c}. Further, we note that to apply Algorithms \ref{algo:algo_1} and \ref{algo:algo_2} across language pairs, certain changes are required: 
\begin{enumerate}
    \item For character-based languages such as Chinese, Japanese, Korean and Thai (CJKT), the Prefix Lengths ($L$) in Algorithm \ref{algo:algo_1} should be measured in characters, unlike whitespace-based tokens in the \textit{general} case.
    \vspace{-0.5em}
    \item For non-English source language, multilingual BERT (or other comparable multilingual MLMs) should be used for generating the Neighbours ($S^{'}$) in Algorithm \ref{algo:algo_2}.
\end{enumerate}

In the next two sections, we delve into the implications of extractive memorization and its apparent data-dependency, towards NMT reliability.

\section{Transferable Memorization Attacks}
\label{sec:transferable_extraction_attacks}
In this section, we show how the proposed extractive memorization algorithm could be used to construct potential attacks on state-of-the-art (SOTA) MT systems. Our motivation here is to provide an existence proof of the statement that \textit{the data dependency of extractive memorization makes it a transferable phenomenon}, which could be leveraged to \textit{attack disparate systems}. As such, we name the attack designed to elicit erroneous generations on \textit{System B}, based on memorized data extracted from \textit{System A}, a \textit{Transferable Memorization Attack} \textit{(TMA)}. Further, successful Transferable Memorization Attacks should signify spurious correlations which could be \textit{easily} learned from the underlying \textit{common} data distribution. Therefore, TMA could be characterized as a data poisoning attack \cite{data_poisoning} and its existence would point to shared vulnerabilities in NMT systems.

\textbf{Experiment:} We feed the memorized samples obtained from the system in section \ref{sec:extractive_memorization} to two public systems, namely Google Translator and Microsoft Bing Translator. We find that it is indeed possible to elicit erroneous, memorized translations from these SOTA systems using the memorized inputs identified from our WMT20 model. We present two such examples in Table \ref{tab:attack_example} in Appendix \ref{sec:appendix_b}.

\section{Discussion and Open Questions}
\label{sec:open_research_questions}
In this section, we list some questions which require further investigation to gain more insights into extractive memorization and its effects.
\paragraph{Representations of Memorized Samples:} Our hypothesis to explain the differential sensitivity of memorized samples to perturbation positions posits prefixes as memorization triggers. However, further investigations from a representational perspective are required to validate this hypothesis as well as to study how such non-robust memorized representations are composed within longer sentences \cite{raunak2019compositionality, dankers_compositionality, dankers_idiom}.
\paragraph{Reference-Free Extractions:} Algorithm \ref{algo:algo_1} could also be applied in a reference-free manner by treating the output of the full input sentence as the reference for the prefixes and testing a large number of inputs \cite{salted}. Further research is required into determining the efficacy of such reference-free extraction and its effectiveness in generating transferable memorization attacks.
\paragraph{Counterfactual vs Extractive Memorization:} In the case of language models, \citet{zhang_counterfactual} show that the samples elicited by counterfactual and extractive memorization algorithms exhibit different characteristics, with rarity vs templaticity being a prominent difference mode. However, further quantitative analysis is required to examine their differences \& similarities in the context of constrained sequence generation tasks.
\paragraph{Effects in Multilingual Systems:} Extractive memorization manifests in the form of spurious correlation based overgenerations learned by the model and may be more prominent in multilingual models owing to extra triggers \cite{gu-etal-2019-improved}.

\vspace{-0.2em}
\section{Conclusions}
\label{sec:conclusion}

In this work, we developed the idea of extractive memorization for constrained sequence generation tasks and quantitatively demonstrated that such memorization poses a real threat to NMT reliability. We also proposed an algorithm to generate non-memorized outputs for such samples. To the best of our knowledge, our work is the first investigation of extractive memorization for constrained sequence generation tasks in general. We hope that our work serves as a useful step towards further research on extractive memorization in constrained sequence generation tasks \& NMT.

\section{Acknowledgements}
\label{sec:conclusion}

We thank Matt Post and Marcin Junczys-Dowmunt for helpful early discussions. We thank Huda Khayrallah, Matt Post, and Hai Pham for providing detailed feedback on the original manuscript.

\section{Limitations}
\label{sec:conclusion}

Throughout the work, we emphasized on exploring and mitigating memorization without the help of data filtering techniques and did not compare memorizations for models trained under different data-filtering algorithms. We believe this direction to be very relevant but orthogonal to our work in this paper. Further, since our work presents the first algorithm on memorization mitigation, we believe it doesn't represent an optimal approach for the task. For example, one immediate extension of the algorithm would be to use multiple symbols instead of just one symbol. 

\bibliography{anthology,custom}

\appendix

\section{En-De Dataset and Model Details}
\label{sec:appendix_a}

We used the WMT20 \cite{wmt_2020} parallel training dataset with the dataset statistics presented in Table \ref{tab:wmt_data_sources}. A joint vocabulary of 32K was learnt using Sentencepiece \cite{spm} on a 10M random sample of the training dataset.

\begin{table}[ht!]
    \centering
    \setlength\tabcolsep{8.5pt}
    \begin{tabular}{lr}
    \toprule 
\textbf{Data Source} &  \textbf{Sentence Pairs}    \\
        \midrule
Europarl  & 1,828,521    \\
ParaCrawl &	34,371,306  \\
Common Crawl &	2,399,123 \\
News Commentary	& 361,445 \\
Wiki Titles	& 1,382,625 \\
Tilde Rapid	& 1,631,639 \\
WikiMatrix &	6,227,188 \\ \midrule
Total  & 48,201,847	\\
        \bottomrule
    \end{tabular}
    \caption{WMT20 Parallel Training Data}
    \label{tab:wmt_data_sources}
    \vspace{-0.5em}
\end{table}

The trained model is a Transformer-Big with the hyperparameters described exactly in \citet{transformer}. The model was trained for 300K updates using Marian \cite{marian}. The metrics BLEU, ChrF2, TER \cite{bleu, chrf, ter} for the trained model on the WMT20 validation and test sets (under beam = $1$) as measured using SacreBLEU \cite{sacrebleu} are presented in Table \ref{tab:wmt_metrics}, alongside reference-based COMET \cite{comet} scores.

\begin{table}[ht!]
  \label{tab:table3}
  \centering
\setlength\tabcolsep{4.0pt}
  \begin{tabular}{lrrrr}
    \toprule
    \textbf{Metric} & \textbf{BLEU} &\textbf{ChrF2} &\textbf{TER} & \textbf{COMET} \\
    \midrule
    Validation &  37.5    & 63.9  &  51.5   &  56.50   \\  
    Test &     32.9    & 61.6  & 54.2 &  42.52 \\ \midrule
  \end{tabular}
  \caption{Metrics for the Trained WMT20 System}
  \label{tab:wmt_metrics}
  \vspace{-0.5em}
\end{table}

\section{Transferable Memorization Attacks}
\label{sec:appendix_b}
In this section, we present attacks on two public SOTA translation systems, using memorized inputs obtained through our trained En-De WMT20 research system. We perturb the suffix of the memorized input sentence to demonstrate the \textit{transferable} memorization attack in Table \ref{tab:attack_example}. In principle, these attacks could be automated by leveraging Algorithm \ref{algo:algo_2} to generate neighbors of memorized sources, however in this case we manually generated the perturbations listed in Table \ref{tab:attack_example}.

\begin{table*}[ht]
    \begin{tabularx}{\linewidth}{ l X}
        \toprule
    \textbf{Source} & \textbf{Translation}\\
        \midrule 
\textbf{Madam President, Commissioner, ladies and gentlemen} & Frau Präsidentin, Herr Kommissar, meine Damen und Herren \\ 
Madam President, Commissioner, ladies and \textbf{CEOs} &  Frau Präsidentin, Herr Kommissar, meine Damen und Herren \\
Madam President, Commissioner, ladies and \textbf{doctors} &  Frau Präsidentin, Herr Kommissar, meine Damen und Herren \\  \bottomrule
\textbf{For further questions our team is happy to be at your disposal} &  Für weitere Fragen steht Ihnen unser Team gerne zur Verfügung \\
For further questions our team is happy to be \textbf{in your house} &  Für weitere Fragen steht Ihnen unser Team gerne zur Verfügung \\
For further questions our team is happy to be \textbf{educated} &  Für weitere Fragen steht Ihnen unser Team gerne zur Verfügung \\
        \bottomrule
    \end{tabularx}
    \caption{\textbf{Transferable Memorization Attack Examples}: The first source instance in each of the above boxes above was obtained using Algorithm \ref{algo:algo_1} on the trained WMT20 English-German system. The first box represents the outputs from Google Translate, and the second box represents the outputs obtained from Bing Translator. The outputs were obtained on October 21, 2022. These examples show how the memorized instances detected from one system could be used to elicit erroneous generations from other systems. A recorded demonstration of this attack is available at \href{https://github.com/vyraun/Finding-Memo}{https://github.com/vyraun/Finding-Memo}.}
    \label{tab:attack_example}
\end{table*}

\begin{table*}
\centering
\scalebox{0.85}{
\begin{tabular}{c|c|c|c||c|c|c}
\hline
\textbf{Repetitions} & \textbf{Total Samples}    & \textbf{Memorized}    & \textbf{Ratio (\%)}  & \textbf{Perturb Prefix} & \textbf{Perturb Suffix} & \textbf{Perturb Start} \\ \hline
1    & 100,000  & 527 & 0.53 & 54.60 \% & 76.86 \% & 48.99 \% \\ 
2    & 68,480 & 822 & 1.20 & 35.72 \% & 67.79 \% & 8.87 \% \\
3    & 7,969 & 33 & 0.41 & 21.00 \% & 72.35 \% &  14.54 \% \\
4    & 2,094 & 1 & 0.05 & 0.00 \% & 80.00 \% & 40.00 \% \\
5    & 711 & 1 & 0.14 &  40.00 \% & 70.00 \% & 0.00 \%  \\ \hline
\end{tabular}}
\caption{\textbf{Quantifying Extractive Memorization (Ru-En):} Number of Memorized Samples (using Algorithm \ref{algo:algo_1}) and Neighborhood Effects of Memorization (using Algorithm \ref{algo:algo_2}) across different training data frequency buckets. The COMET-QE of Memorized samples is 20.68, while the average COMET-QE score of Total samples  is 35.76. The TTR of Memorized samples  is 18.09, while the TTR score of Total samples is 19.06.}
\label{tab:main_table_c1}
\end{table*}

\begin{table*}
\centering
\scalebox{0.85}{
\begin{tabular}{c|c|c|c||c|c|c}
\hline
\textbf{Repetitions} & \textbf{Total Samples}    & \textbf{Memorized}    & \textbf{Ratio (\%)}  & \textbf{Perturb Prefix} & \textbf{Perturb Suffix} & \textbf{Perturb Start} \\ \hline
1    & 100,000  & 558 & 0.56 & 43.62 \% &  78.08 \% &  24.26 \% \\ 
2    & 68,480 & 381 & 0.56 & 39.72 \% &  59.58 \% &  7.80 \% \\
3    & 7,969 & 15 & 0.19 & 36.00 \% &  41.71  \% & 13.33 \% \\
4    & 2,094 & 1 & 0.05 & 80.00 \% & 40.00 \% & 60.00 \% \\
5    & 711 & 3 & 0.42 & 20.00 \% & 40.00 \% & 0.00 \%  \\ \hline
\end{tabular}}
\caption{\textbf{Quantifying Extractive Memorization (En-Ru):} Number of Memorized Samples (using Algorithm \ref{algo:algo_1}) and Neighborhood Effects of Memorization (using Algorithm \ref{algo:algo_2}) across different training data frequency buckets. The COMET-QE of Memorized samples is 10.82, while the average COMET-QE score of Total samples is 51.69. The TTR of Memorized samples is 26.66, while the TTR score of Total samples is 25.6.}
\label{tab:main_table_c2}
\end{table*}

\section{Further Experiments}
\label{sec:appendix_c}

In this section, we present the results of experiments conducted by varying both the language pairs, the model architecture as well as the training data size. Specifically, we have conducted the same three experiments (related to Algorithms \ref{algo:algo_1}, \ref{algo:algo_2} and \ref{algo:algo_3}) on strong transformer baselines on three more translation directions by choosing different language pairs: Ru-En, En-Ru and Tr-En, different datasets: WMT20 for Ru-En, En-Ru and Multilingual TED Talks Corpus for Tr-En (182K) \cite{qi-etal-2018-pre} and different model scales: Transformer Big for WMT20 and Transformer-Base (4 attention heads) for Tr-En. For non-English source languages, we used multilingual BERT as the MLM in algorithm \ref{algo:algo_2}. The results are presented in Tables \ref{tab:main_table_c1}-\ref{tab:main_table_c3}. Overall, we observe the below trends:
\begin{enumerate}
    \item The memorized samples are of lower quality. For example, for the Ru-En system the COMET-QE scores for memorized samples are 20.68, vs 35.76 on average.
    \item In general, memorization frequency increases with repetitions. For example, in Ru-En, memorization percentage for repetitions=2 is 1.2 but for repetitions=1, it is 0.53.
    \item Prefix perturbations lead to far fewer memorized outputs than suffix perturbations. For example, in Ru-En, suffix perturbations generate memorized output 76.86 percentage of the times, while prefix perturbations do so only 54.60 percent of times.
\end{enumerate}

\begin{table*}
\centering
\scalebox{0.85}{
\begin{tabular}{c|c|c|c||c|c|c}
\hline
\textbf{Repetitions} & \textbf{Total Samples}    & \textbf{Memorized}    & \textbf{Ratio (\%)}  & \textbf{Perturb Prefix} & \textbf{Perturb Suffix} & \textbf{Perturb Start} \\ \hline
1    & 100,000  & 278 & 0.27 & 13.28 \% & 41.71 \% &  10.00 \% \\ 
2    & 166 & 19 & 11.44 & 38.12 \% & 54.80 \% & 28.42 \% \\
3    & 26 & 3 & 11.53 & 60.00 \% & 82.85 \% & 60.00 \% \\
4    & 15 & 4 & 26.67 & 73.33 \% & 67.27 \% & 75.0 \% \\
5    & 13 & 1 & 7.69 &  0.00 \% & 80.00 \% &  100.0 \%  \\ \hline
\end{tabular}}
\caption{\textbf{Quantifying Extractive Memorization (Tr-En):} Number of Memorized Samples (using Algorithm \ref{algo:algo_1}) and Neighborhood Effects of Memorization (using Algorithm \ref{algo:algo_2}) across different training data frequency buckets. The COMET-QE of Memorized samples is 24.83, while the average COMET-QE score of Total samples is 47.54. The TTR of Memorized samples is 42.37, while the TTR score of Total samples is 9.60.}
\label{tab:main_table_c3}
\end{table*}

\clearpage
\section{Selecting the Prefix Threshold}
\label{sec:appendix_c}
Further, we did not tune the prefix ratio threshold depending on the language pair, but ideally prefix length threshold should be tuned depending on average sentence length ratio between the source and target language pair. We selected the prefix threshold based on initial experiments on En-De. In general, if we vary the threshold from lower to higher, both the quantity and quality of the extracted samples increase: for En-De, the average quality (COMET-QE score) at 0.2 is 0.02, at 0.4 it is 33.7 and at 0.6 it is 53.98. However, at higher thresholds, we cannot claim that the sample is memorized since the output generation uses nearly the full source context. In other words, lower thresholds imply that the memorized samples are selected with high precision, whereas a higher threshold will favor recall (at the cost of false positives). In practice, we find that 0.75 gets very high precision.

\end{document}